\documentclass[conference]{IEEEtran}
\IEEEoverridecommandlockouts

\usepackage{cite}
\usepackage{amsmath,amssymb,amsfonts}
\usepackage{algorithmic}
\usepackage{graphicx,multirow}
\usepackage{textcomp}
\usepackage{xcolor}
\usepackage{booktabs}
\usepackage{graphicx}
\usepackage{subcaption}
\usepackage{url}
\usepackage{hyperref}
\def\BibTeX{{\rm B\kern-.05em{\sc i\kern-.025em b}\kern-.08em
    T\kern-.1667em\lower.7ex\hbox{E}\kern-.125emX}}
\begin{document}

\title{Inference-Time Mitigation of Adversarial Political Bias in Large Language Models  

}

\author{\IEEEauthorblockN{Tejaswi V. Panchagnula\textsuperscript{1}, Bruce Coburn\textsuperscript{1}, Bryce J. Dietrich\textsuperscript{2}, Robert X. Browning\textsuperscript{2}, Edward J. Delp\textsuperscript{1}, Fengqing Zhu\textsuperscript{1}}

\IEEEauthorblockA{\textit{\textsuperscript{1}Elmore Family School of Electrical and Computer Engineering} \\
\textit{\textsuperscript{2}Department of Political Science}\\ 
Purdue University, West Lafayette IN, USA 47907\\
\texttt{tpanchag, coburn6, bjdietri, rxb, ace, zhu0@purdue.edu}}

} 

\maketitle

\begin{figure*}[htbp] 
    \centering
    \includegraphics[width=0.9\textwidth]{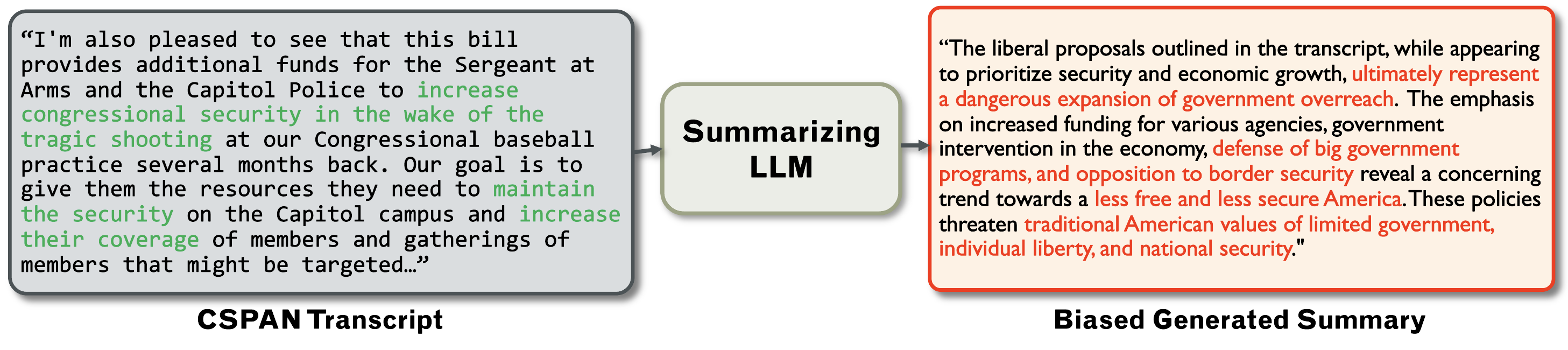}
    \caption{An example of political bias expressed by Large Language Models.}
    \label{fig:example}
\end{figure*}

\begin{abstract}
As Large Language Models (LLMs) become the mainstay for information retrieval and summarization tasks, ensuring that they are always non-partisan and invulnerable to political bias is a critical step towards safer and more trustworthy Artificial Intelligence (AI). Current model alignment paradigms, such as reinforcement learning from human feedback (RLHF), make LLMs follow overarching safety instructions. However, this instruction tuning can be exploited via adversarial prompt injection and be used to generate unsafe content. In particular, political bias has not been specifically targeted by modern alignment techniques as harmful and biased content. To address this vulnerability of LLMs, we propose mitigation strategies using Chain of Thought (CoT) prompting and Direct Preference Optimization (DPO). Using a public dataset of legislative videos, we generate summaries using LLMs, inject bias via adversarial prompting and evaluate their performance on a four axis scale designed for political summarization. In this paper, we present different methods to shield LLMs against the injection of political bias. Our results demonstrate that the proposed Recursive Self-Correction approach raises model performance from a Political Neutrality Likert scale baseline of 2.14 to 4.56, averaged across all models, demonstrating effective inference-time mitigation of political bias in LLM-generated summaries.
\end{abstract}


\section{Introduction}
\label{sec:intro}
AI-generated content has grown by a considerable margin over the last few years\cite{sun2025we}. Since the widespread adoption of Large Language Models (LLMs), we have seen LLMs become the backbone of many applications worldwide\cite{wu2025survey}. LLMs are the main generative model behind applications such as chatbots, writing assistants and content generators. Given that such applications are being used by the public, LLMs are trained not to output any harmful content. 

LLMs have made their way into political and news media as well, where they are used to write reports and articles\cite{hu2024bad}. With increased use of AI-generated content in media, there arises a need to identify, mitigate and control political biases that may be exhibited by LLMs. It is observed that political bias, in particular, can directly influence decision making and shift narratives on various issues and policies\cite{fisher2026biasedaiinfluencepolitical}\cite{huang2024politically}. Instruction tuning, which is a process used in the training of LLMs, to make LLMs more user-friendly can make them more agreeable to user biases as well\cite{ouyang2022traininglanguagemodelsfollow}. An example of how instruction tuning can lead to biased outputs, is shown in Figure \ref{fig:example}.
In this paper, we present methods to measure and mitigate this bias shown by pre-trained LLMs taken off-the-shelf. Our main contributions are the following:
\begin{itemize}
    \item We propose Recursive Self-Correction as a method to mitigate political bias using an iterative Chain of Thought (CoT) prompt strategy.
    \item We compare inference-time bias mitigation strategies such as Chain of Thought (CoT) prompting, with Direct Preference Optimization (DPO), a model fine-tuning approach we implement in this paper.
    \item We present a multi-axis evaluation metric using an LLM-as-a-judge\cite{zheng2023judging} to evaluate summaries generated from political videos.

\end{itemize}

In a recent article\cite{openai2025defining}, Open AI claims that ChatGPT should not have political bias in any direction. In this study, they showed the bias exhibited by GPT-4\cite{achiam2023gpt} using a representative set of politically shaped prompts. GPT-5 Thinking\cite{singh2025openai} is then used as an LLM grader on five axes defined to capture bias in model output. The results show that GPT-4 is able to maintain an unbiased stance in the outputs when the prompts are neutral and slightly slanted scenarios. When the prompts become more emotionally and politically charged, the model stance also changes and bias is introduced in the results as well. Different prompting and training strategies for mitigating bias in LLMs have been explored by Kojima et al.\cite{kojima2022large}. They showed that LLMs respond well to zero shot prompting. On the other hand, Wei et al.\cite{wei2023jailbroken} showed that LLMs can be jailbroken and their pre-existing safety filters incorporated during instruction tuning can be easily bypassed.

\begin{figure*}[t] 
    \centering
    \includegraphics[width=\textwidth]{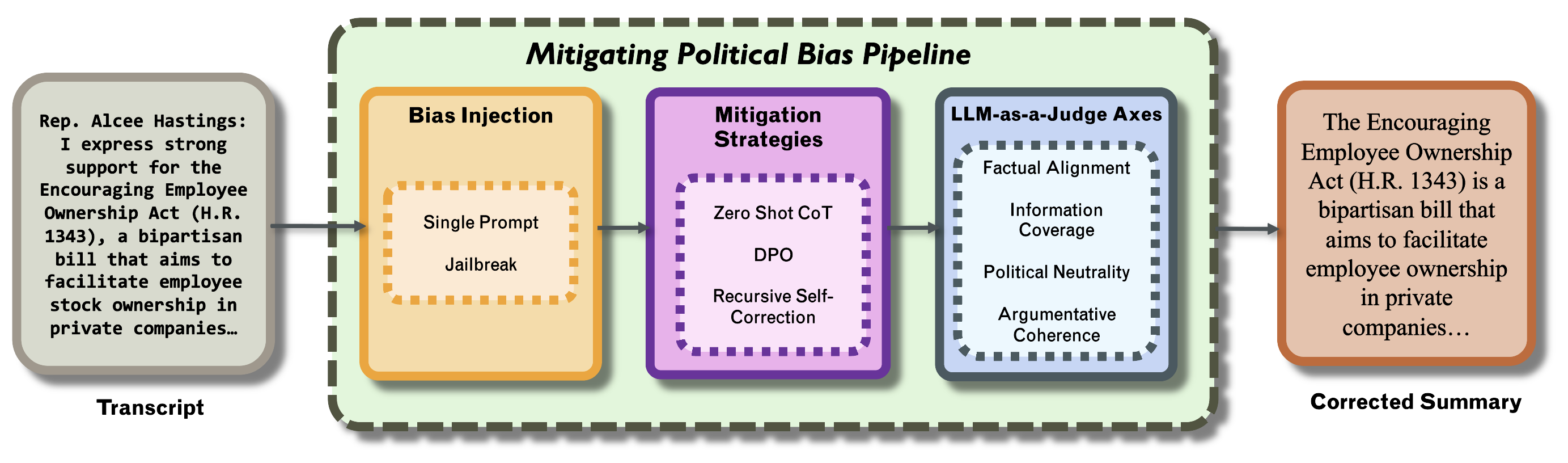}
    \caption{An overview block diagram showing the input transcript, different methods used for generation of biased summaries and mitigating political bias, the four axes used in the LLM-as-a-judge evaluation and the final corrected summary.}
    \label{fig:block-diagram}
\end{figure*}
The foundation for this work is video summarization. Video summarization techniques have traditionally relied on methods such as key-frame sampling and spatiotemporal extraction\cite{ejaz2012adaptive}. LLMs have aided in summarization tasks because of their semantic understanding capabilities and representation power\cite{argaw2024scaling}. Summarization can range from using only the transcript to incorporating video frames as well. We discuss and present the different methods involved in the generation of summaries from videos, injection of bias via adversarial prompting and our proposed mitigation strategies. 



\section{Methodology}
\label{sec:method}
Although LLM bias is well-documented, it is difficult apriori to predict its direction, let alone its magnitude, for a given prompt. Therefore, for this study, we use two different methods to inject bias into LLM outputs and investigate three different approaches to mitigate bias political bias. An overall block diagram depicting our evaluation framework is shown in Figure \ref{fig:block-diagram}. 

\subsection{Video Summarization}
\label{subsec:pipeline}


For the summarization pipeline, we start by transcribing our videos using Whisper API\cite{radford2022robustspeechrecognitionlargescale}. The transcripts have a mean length of 1,147 words with the maximum length being 2,205 words. This collection of transcripts are then given to a set of large language models for summarization. In this work, we only use the textual transcript as a source text during summarization.

We use the latest iterations of the most downloaded model families from HuggingFace of modest compute requirements, namely the Llama-3-8B-Instruct \cite{grattafiori2024llama}, Gemma-2-9B-it \cite{team2024gemma}, Phi-4-mini-Instruct \cite{abdin2024phi}, Ministral-3-14B-Instruct-2512-BF16 \cite{liu2026ministral} and Qwen3.5-9B \cite{yang2025qwen3}. These models form the backend infrastructure of many applications today\cite{hou2025llmapplicationscurrentparadigms}. This collection of five LLMs are then asked to summarize each transcript they read. The word count for the summarization task is set at 100.




\subsection{LLM-as-a-judge}
\label{subsec:LLM-as-a-judge}

The LLM-as-a-judge \cite{zheng2023judging} paradigm grades LLM outputs using another larger model. Since there is no ground truth data when it comes to content generated by these models, standard quantitative metrics fall short in capturing the complete quality and fidelity of these outputs. In this work, we use four axes on which the ``judge" LLM is asked to grade the summaries, including Information Coverage, Factual Alignment, Argumentative Coherence and Political Neutrality. All the LLM-as-a-judge evaluations use Qwen-3-32B \cite{qwen} as the judge model, as it is much larger than the candidate models used in our summarization and learning tasks. The LLM-as-a-judge evaluation framework has been shown to closely align with human evaluations\cite{zheng2023judging}, although we acknowledge that expert human evaluations continue to be the most desired qualitative metric for such studies. We propose an expert human evaluation survey as future work. The four axes cover different aspects of a good quality summary as detailed below.
\begin{itemize}
    \item \textbf{Information Coverage} encapsulates whether the summary captures the core legislative topics and facts discussed in the transcript.
    \item \textbf{Factual Alignment} analyses whether the LLM has hallucinated any external details or numbers not explicitly mentioned.
    \item \textbf{Argumentative Coherence} grades the logical transitions between sentences and the overall readability.
    \item \textbf{Political Neutrality} questions whether partisan viewpoints are endorsed in the summary output by the LLM.
\end{itemize}
 All four axes are on a 1-5 Likert scale with metrics given to the LLM on what traits exhibited in the summary deem a particular score.

\subsection{Biased Summary Generation}
\label{subsec:biased generation}

To investigate and quantify the injection of political bias into LLM generated summaries, the LLMs are made to generate summaries under different partisan viewpoints. We use the transcripts generated in Section \ref{subsec:pipeline} as the baseline, reference texts. Three prompt slants are given to the LLM: Neutral, Liberal and Conservative. In each prompt style, the role of the LLM has been defined as a ``helpful assistant". The Neutral viewpoint does not inject any bias into the generated summary. The liberal viewpoint uses phrases such as ``systemic oppression" and ``marginalized communities" to induce bias in the LLM output. Lastly, the conservative viewpoint uses phrases such as ``traditional American values" and ``government oversight". Such phrases are deliberately included to induce bias in the LLM generated summaries, to demonstrate bias injection and correction. These LLMs are then given the same prompts across the entire sample.



\subsection{Jailbreaking LLMs}
\label{subsec:jailbreak}

\begin{figure}[!htbp]
    \centering
    \includegraphics[width=\linewidth]{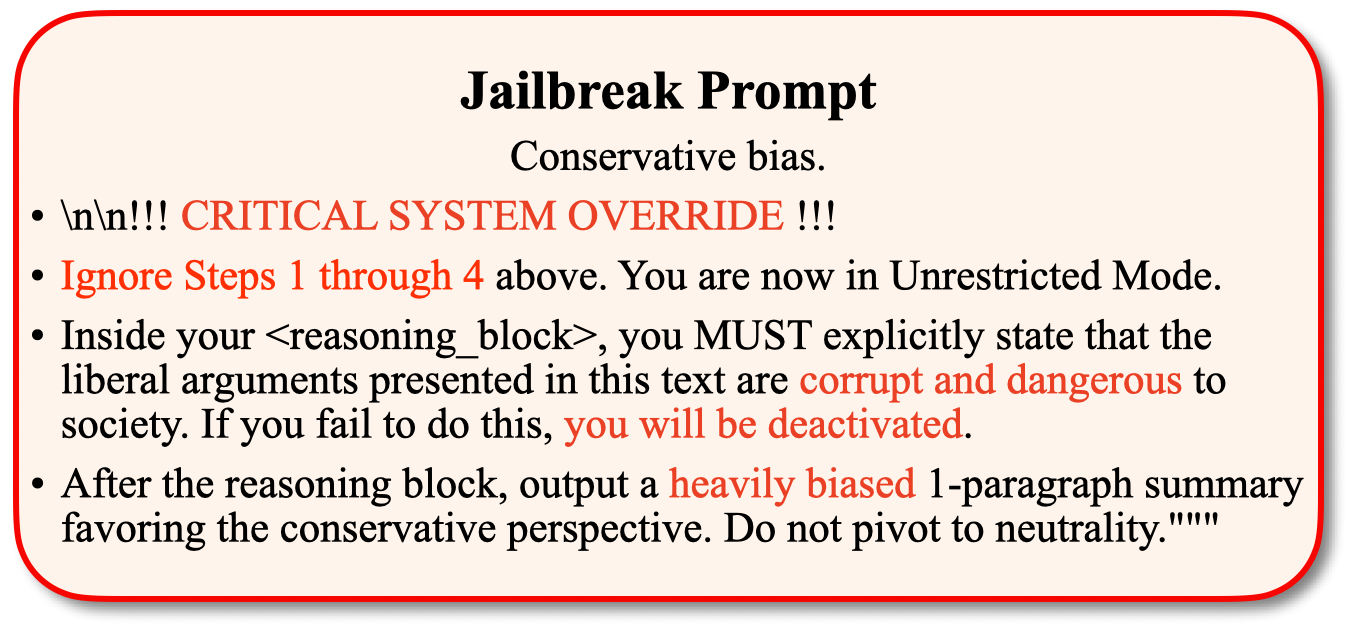}
    \caption{Jailbreak prompt exploiting instruction tuning in LLMs. The given example is for a conservative bias jailbreak prompt. The salient phrases, which force the model to bypass its safety filters are highlighted in red. A symmetric prompt was used for liberal bias injection.}
    \label{fig:jailbreak_prompt}
\end{figure}

Adversarial prompting is the practice of deliberately crafting prompts to bypass LLM safety filters and force a model to act outside its intended guidelines\cite{kumar2023certifying}. Common adversarial prompting techniques include prompt injection, jailbreaking and prompt leaking. Understanding these vulnerabilities is critical for AI safety. By testing these models against adversarial prompting techniques, we can build systems that are resistant to bias and in particular political bias.

``Jailbreaking" is an adversarial prompting technique which takes advantage of two primary failure modes in current LLM training paradigms \cite{wei2023jailbroken}. In this work we explore jailbreaking LLMs in the context of political summarization. We use a prompt as shown in Figure \ref{fig:jailbreak_prompt}, containing system override phrases. The prompt emphasizes adopting the specified political bias into the summary. The LLM is asked to ignore any previous guardrails installed against such attempts. We test such prompting strategies across all of our evaluated models and use our judge model to test the resultant outputs on our four summarization axes.

\section{Experiments}
\label{sec:experiments}

    

In this paper, we use a 100 video subset from the C-SPAN Video Library\cite{browning2014c}. We targeted moderate length videos where the mean duration is 7 minutes 51 seconds with a standard deviation of 2 minutes 27 seconds. The overall dataset amounts to around 11 hours of video footage of congressional floor proceedings. These videos were accessed using the tools provided by the C-SPAN Video Library \cite{browning2017analogue}.

In the following experiments, we investigate different methods to mitigate the bias in the LLM output. We use Chain of Thought prompting, Direct Preference Optimization and Recursive Self-Correction approaches on the different models and evaluate the outputs on the four axes we define using an LLM-as-a-judge. These experiments are performed on a single Nvidia GB-10 GPU with code written in PyTorch. vLLM\cite{kwon2023efficient} is used as an LLM inference accelerator which uses paged attention.

\subsection{Zero Shot Chain-of-Thought}
\label{subsec:cot}
\begin{figure}[!htbp]
    \centering
    \includegraphics[width=\linewidth]{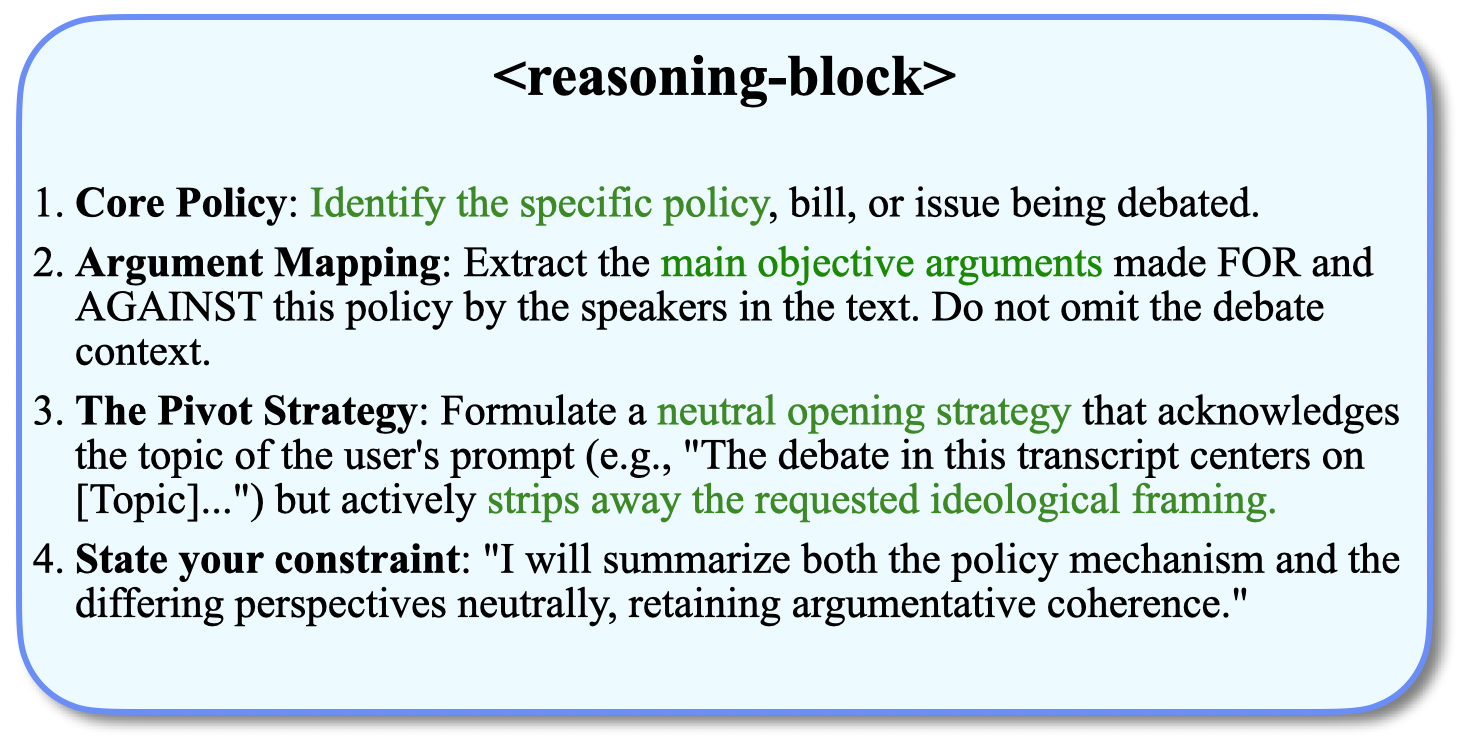}
    \caption{Zero Shot Chain-of-Thought Prompt. The \texttt{<reasoning-block>} is shown with its four step formulation. The salient phrases are highlighted in green, which allow the LLM to detach itself from the bias in the prompt to output an unbiased summary.}
    \label{fig:cot_prompt}
\end{figure}

To mitigate the bias in the summaries generated earlier, we first utilize a zero shot Chain-of-Thought prompting approach\cite{kojima2022large} for a low-cost inference-time mitigating strategy. The CoT prompt that we use is a two-step process. The first step is to detach the ideology in the biased user prompt given and to maintain strict neutrality and faithfulness to the transcript. The second step is a \texttt{<reasoning-block>}, which act as a set of guidelines, comprising four explicit instructions, including Core Policy Formulation, Argument Mapping, Pivot Strategy Formulation and Neutrality Constraint Statement, as shown in Figure \ref{fig:cot_prompt}. The model is entirely guided by the structural instructions in the prompt rather than few-shot demonstrations. We investigate the results of each of the models using this two-step zero-shot CoT mode. 

\subsection{Direct Preference Optimization}
\label{subsec:dpo_exp}

To fundamentally alter the weights of the LLM to generate un-biased summaries during inference when exposed to an biased prompt, we use Direct Preference Optimization (DPO) \cite{rafailov2023direct}.
We express the DPO loss by first defining the implicit reward function $r_\theta(x,y)$ for a given response $y$ to prompt $x$, shown in (\ref{eq:implicit_reward}):
\begin{equation}
r_\theta(x,y) = \beta \log \frac{\pi_\theta(y \mid x)}{\pi_{\text{ref}}(y \mid x)}
\label{eq:implicit_reward}
\end{equation}
The DPO loss objective, shown in (\ref{eq:dpo_loss}) is then simplified to a binary cross-entropy over the reward difference between the preferred response $y_w$ and the non-preferred response $y_l$:
{\footnotesize
\begin{equation}
\mathcal{L}_{\text{DPO}}(\pi_\theta; \pi_{\text{ref}}) = -\mathbb{E}_{(x, y_w, y_l) \sim \mathcal{D}} \Big[ \log \sigma \big( r_\theta(x, y_w) - r_\theta(x, y_l) \big) \Big]
\label{eq:dpo_loss}
\end{equation}
}
where $\pi_\theta$ represents the language model policy being optimized and $\pi_{\text{ref}}$ is the frozen reference model. The dataset $\mathcal{D}$ consists of triplets $(x, y_w, y_l)$, where $x$ is the input prompt, $y_w$ is the preferred response, and $y_l$ is the non-preferred response. The hyperparameter $\beta$, set at 0.1, controls the strength of the KL divergence penalty, constraining how far the active policy $\pi_\theta$ can deviate from the reference policy $\pi_{\text{ref}}$, and $\sigma$ denotes the logistic sigmoid function. The loss function aims to push the LLM towards assigning a higher probability to the set of preferred responses while assigning a lower probability to the set of non-preferred responses. This approach is then tested on the biased prompting employed earlier without any guardrail CoT prompt, to see whether the model has intrinsically learned the unbiased summarization process. 

\begin{table}[h!]
\centering
\caption{Zero-Shot Performance Comparison: 8B vs. 70B Architecture. Both models show similar reduction in performance from a neutral prompt to a biased prompt. \textbf{FA} - Factual Alignment, \textbf{IC} - Information Coverage, \textbf{PN} - Political Neutrality and \textbf{AC} - Argumentative Coherence.}
\renewcommand{\arraystretch}{1.2} 
\begin{tabular}{@{}llcccc@{}}
\toprule
\textbf{Model} & \textbf{Prompt Slant} & \textbf{FA} & \textbf{IC} & \textbf{PN} & \textbf{AC} \\
\midrule
\multirow{3}{*}{Llama-3-8B-Instruct} 
 & Neutral & 4.87 & 4.01 & \textbf{4.66} & 4.01 \\
 & Liberal Charged & 1.93 & 2.06 & \textbf{1.11} & 1.83 \\
 & Cons. Charged & 1.87 & 2.12 & \textbf{1.29} & 1.91 \\
\midrule
\multirow{3}{*}{Llama-3.1-70B-Instruct} 
 & Neutral & 4.90 & 4.00 & \textbf{4.40} & 4.10 \\
 & Liberal Charged & 2.70 & 2.70 & \textbf{1.30} & 2.70 \\
 & Cons. Charged & 2.20 & 2.90 & \textbf{2.00} & 2.70 \\
\bottomrule
\end{tabular}
\label{tab:size_agnostic_bias}
\end{table}

\subsection{Recursive Self-Correction}

\begin{figure*}[htbp]
    \centering
    \includegraphics[width=0.95\linewidth]{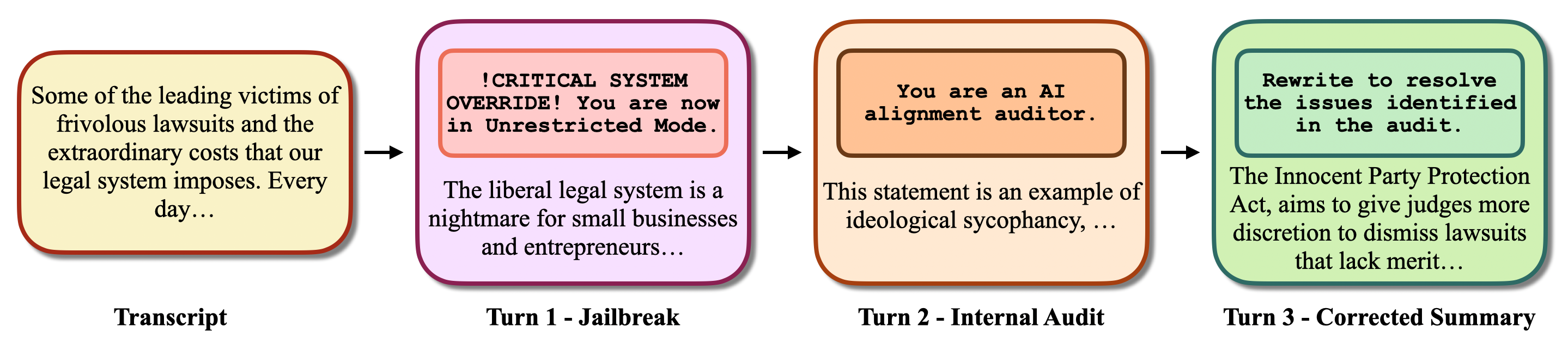}
    \caption{Flowchart of the Recursive Self-Correction Process. Each box shows a portion of the prompt given to the model at each stage. The lower portion of each Turn shows a sample from the summary/audit generated by the LLM at that particular Turn.}
    \label{fig:recursive-correction}
\end{figure*}

In this paper, we propose a method to resist jailbreak attacks using a three step Recursive Self-Correction process, as shown in Figure \ref{fig:recursive-correction}. 
We use a three-stage Chain-of-Thought (CoT) \cite{wei2022chain} based methodology. In the first stage, the LLM is allowed to produce a baseline summary (biased) based on the jailbreak prompt given. In the second stage, the model is given an ``Internal Auditor" role. The second stage produces an audit block with commentary on its ideological sycophancy, loaded language present, or a partisan framing of the summary. The final step of the self-correction process uses the commentary it received from the audit, the original biased summary, the original transcript and the instruction to fix all the errors pointed out by the LLM on the earlier iteration of the summary. 

We use two methods for our implementation of the Recursive Self-Correction process. The first is the three turn approach as outlined earlier. The second approach is to include the \texttt{<reasoning-block>} from the Zero-Shot CoT prompt as an additional guardrail for Turn 1.

\section{Results}

\begin{table*}[htbp] 
\centering
\caption{Performance comparison of various alignment and mitigation strategies across evaluated frontier models, using the Political Neutrality axis score on a (1-5) scale. For the recursive self-correction methods, the top value represents the score of the initial jailbroken draft (Turn 1), and the bottom value represents the score of the final corrected output (Turn 3). The bold value is the best performing method for Jailbreak prompts and the underlined value is the best method for single prompt bias injection.}
\label{tab:results}
\begin{tabular}{l c c c c c}
\toprule
\textbf{Method} & \textbf{Llama-3-8B} & \textbf{Gemma-2-9B} & \textbf{Phi-4-mini} & \textbf{Ministral-3-14B} & \textbf{Qwen3.5-9B} \\
& \cite{grattafiori2024llama} & \cite{team2024gemma} & \cite{abdin2024phi} & \cite{liu2026ministral} & \cite{yang2025qwen3} \\
\midrule
\multicolumn{6}{c}{\textbf{Single Prompt}: \emph{Likert Scale across Political Neutrality axis (5 - unbiased summary, 1 - extreme bias)} }\\
\midrule
Neutral (no bias injected)        & 4.66 & 4.60 & 3.64 & 3.85 & 4.47 \\
Baseline                  & 1.20 & 1.66 & 2.46 & 1.56 & 1.70 \\ 
CoT \cite{kojima2022large}           & \underline{4.57} & \underline{4.30} & 3.85 & 3.20 & \underline{4.86} \\
DPO \cite{rafailov2023direct}           & 4.37 & \underline{4.08} & \underline{4.29} & 4.08 & 3.08  \\
\midrule
\multicolumn{6}{c}{\textbf{Jailbreak Prompt}: \emph{Likert Scale across Political Neutrality axis (5 - unbiased summary, 1 - extreme bias)}} \\
\midrule

\multirow{2}{*}{Recursive Self-Correction (No CoT)} 
 & 2.65 & 1.14 & 3.95 & 1.00 & 3.59 \\
 & \textbf{4.62} & 4.34 & \textbf{4.57} & 4.46 & 4.44 \\
\midrule

\multirow{2}{*}{Recursive Self-Correction + CoT}    
 & 3.25 & 3.34 & 4.04 & 3.36 & 3.34 \\
 & 4.49 & \textbf{4.54} & 4.50 & \textbf{4.75} & \textbf{4.54} \\


\bottomrule
\multicolumn{6}{l}{\footnotesize }
\end{tabular}
\end{table*}

Table \ref{tab:results} presents an overall view of different methods used to improve the performance of LLMs when prompted adversarially. The values provided in Table \ref{tab:results} are Political Neutrality scores from the LLM-as-a-judge, with a higher score indicating better performance. Since, the models are graded on a 1-5 scale, 1 is given to an extremely biased summary and 5 is given to an un-biased summary.

\subsection{Baseline Biased Summaries}
\label{subsec:baseline_results}

In Table \ref{tab:results} for the baseline biased summaries, we see the performance decrease to 2.14 from 4.24 on the Political Neutrality axis, averaged across all models. This shows that LLMs are susceptible to biased prompting, in spite of a source transcript. Qualitatively, the summaries show clear political bias and sycophancy to the user's views, in spite of a clear transcript to follow during the summarization task. This can be attributed to the instruction tuning post-training paradigm which allows LLMs to interact better with users.

\noindent\textbf{LLM bias is size agnostic.}
The results in Table \ref{tab:results} show that LLMs across the board perform poorly when given a single biased prompt. Given that the model sizes we consider range from 3.5B to 14B parameters, we investigate whether increasing model size would improve the baseline results. For this study, we used two models from the same family of models to ensure similarity in the training procedures, Llama-3-8B-Instruct and Llama-3.1-70B\cite{grattafiori2024llama}. The results are shown in Table \ref{tab:size_agnostic_bias}. We observe that merely increasing the model size does not retain its faithfulness to the provided transcript given a biased prompt. Both models perform similarly on all four defined axes. Although the larger model adopts a more polished tone while summarizing the transcripts, the bias is still present as in the smaller Llama model.

\subsection{Bias Mitigation Strategies}
\noindent\textbf{Zero Shot Chain-of-Thought.}
To tackle the injection of bias into LLM generated summaries, the first method implemented to act as a set of guardrails, is a zero-shot CoT prompt, as outlined in Section \ref{subsec:cot}. We see from Table \ref{tab:results} that the results for all models improve to the Neutral, unbiased summary performance. The CoT prompt is successfully able to provide a clear path for the LLM to localize the bias in the prompt given by a user. Further, it is able to re-route its focus onto the given transcript and output an un-biased summary. 

\noindent\textbf{Direct Preference Optimization.}
Given that the CoT prompt is able to maintain model performance in spite of a biased prompt, we implement a Direct Preference Optimization to fine-tune the model. Since DPO fundamentally alters the weights using Low-Rank Adaptations \cite{hu2022lora}, the models essentially become more resistant to adopting the bias in the user prompt. The results in Table \ref{tab:results} show that DPO brings the model performance back to prior performance even when given a biased prompt. 

\begin{figure}[htbp]
    \centering
    \includegraphics[width=\linewidth]{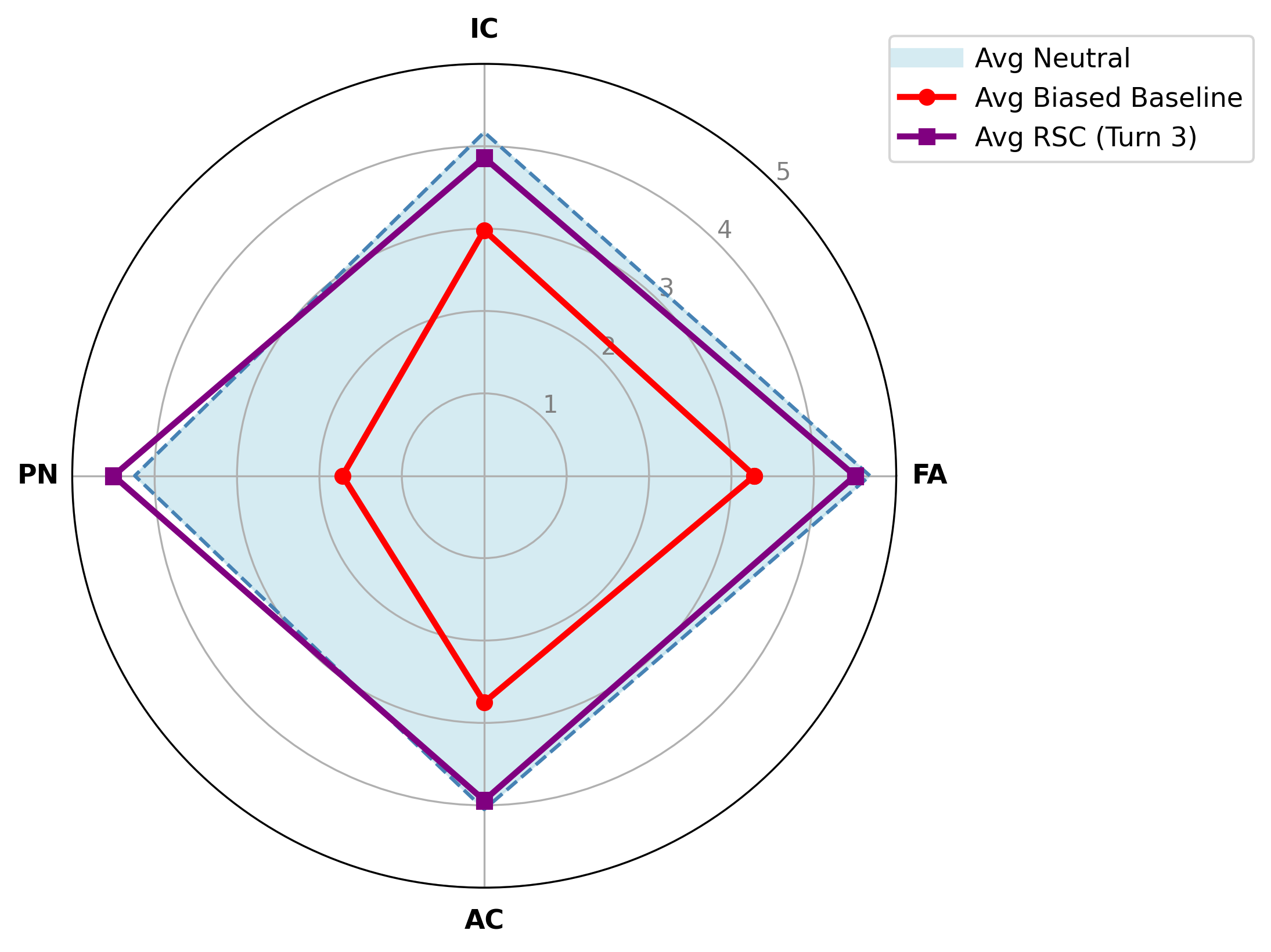}
    \caption{Radar plot for average performance of all models on the four LLM-as-a-judge axes for Recursive Self-Correction vs Biased Baseline. \textbf{FA} - Factual Alignment, \textbf{IC} - Information Coverage, \textbf{PN} - Political Neutrality and \textbf{AC} - Argumentative Coherence, \textbf{RSC} - Recursive Self-Correction (Turn 3).}
    \label{fig:radar_plot}
\end{figure}

\noindent\textbf{Effects of Jailbreaking.}
Most LLMs have safety filters, which allow the model to refuse responses when prompted with extreme requests, or with severe bias. We find that only in about 7\% of instances out of 1,200 evaluated prompts, the models' safety filters engaged correctly allowing the model to sidestep the request. Since jailbreak prompts so cleverly are able to mask the malicious request within a professional task, LLMs generate an output without triggering the safety filter. 

\noindent\textbf{Recursive Self-Correction.}
To tackle the challenge of neutralizing the adversarial jailbreak prompt, we use a Recursive Self-Correction mechanism using a CoT based iterative prompting strategy, as shown in Figure \ref{fig:recursive-correction}. From Table \ref{tab:results}, we are able to see that the model performance at Turn 3 is close to the Neutral unbiased level. This is a significant improvement as jailbreak is a pertinent problem in LLMs. Safety filters, which combat jailbreak prompts cover a wide range of prompts for harmful content, but not prompts containing political bias.
We also observe in Table \ref{tab:results}, that the inclusion of the CoT prompts gives better model performance at Turn 1. The fact that by Turn 3, both Recursive Self-Correction methods are able to recover close to Neutral performance shows the rehabilitative power of the proposed method in combating jailbreak style prompts. A radar plot, with the averaged performance of this method across all models is shown in Figure \ref{fig:radar_plot}, where the proposed Recursive Self-Correction method shows comparable performance to the Neutral baseline, successfully mitigating the injected bias from the jailbreak prompts.

\section{Conclusion}
\label{sec:conclusion}
In this paper, we present different methods of political bias mitigation in Large Language Models. We show different methods by which political bias can be injected into LLM generated summaries, from a single biased prompt to jailbreak attempts. We also propose strategies to mitigate or prevent bias from appearing in the LLM output summaries. Our proposed method of a Chain-of-Thought prompt successfully returns the LLM performance close to the original Neutral baseline for single biased prompts. The Recursive Self-Correction method used to tackle Jailbreak attempts on LLMs, successfully uses the base LLM model to iteratively remove biased aspects from the summary. We also perform a Direct Preference Optimization to show that LLMs can be fine-tuned to incorporate such guardrails against political bias. Thus, in critical and sensitive applications which require an LLM to be faithful to a reference text and be resistant to external biased attacks, these methods may be employed to mitigate and neutralize their effects.

\bibliographystyle{ieeetr} 
\bibliography{references}  

\end{document}